\documentclass[a4paper,11pt]{article}
\usepackage{makecell}
\usepackage{comment}

\usepackage{amsmath}
\usepackage{amsfonts}
\usepackage[flushleft]{threeparttable}
\usepackage{graphicx}
\usepackage[tight,footnotesize]{subfigure}
\usepackage{multirow}
\usepackage{color}
\usepackage{url}
\usepackage[affil-it]{authblk}
\usepackage{wrapfig}

\usepackage[T1]{fontenc}
\usepackage{times}

\newcommand{\myparagraph}[1]{\vspace{4pt}\noindent{\bf #1}}

\let\OLDthebibliography\thebibliography
\renewcommand\thebibliography[1]{
\OLDthebibliography{#1}
\setlength{\parskip}{0.1pt}
\setlength{\itemsep}{0.1pt plus 0.3ex}
}

\begin{document}

\title{Multi-Head Self-Attention via\\Vision Transformer for Zero-Shot Learning}

\author{Faisal Alamri and Anjan Dutta}
\affil{Department of Computer Science, University of Exeter, United Kingdom}
\date{}
\maketitle
\thispagestyle{empty}

\begin{abstract}
Zero-Shot Learning (ZSL) aims to recognise unseen object classes, which are not observed during the training phase. The existing body of works on ZSL mostly relies on pretrained visual features and lacks the explicit attribute localisation mechanism on images. In this work, we propose an attention-based model in the problem settings of ZSL to learn attributes useful for unseen class recognition. Our method uses an attention mechanism adapted from Vision Transformer to capture and learn discriminative attributes by splitting images into small patches. We conduct experiments on three popular ZSL benchmarks (i.e., AWA2, CUB and SUN) and set new state-of-the-art harmonic mean results {on all the three datasets}, which illustrate the effectiveness of our proposed method.
\end{abstract}

% and that it significantly achieves state-of-the-art results. 

% The vital insight is the knowledge transfer, where the knowledge learned during training from the seen classes is transferred to the unseen classes with class embeddings that capture the similarities between the two sets of classes.

%\anjan{This is an anonymous submission. Please do not put link to the code, where our identities are revealed. You don't need to put  link to the code at the moment.}
%The source code is released at \url{https://github.com/Faisal-Anjan/VTA-GSZL}


\begin{wrapfigure}{r}{0.45\textwidth} 
\vspace{-30pt}
\begin{center}
\includegraphics[width=0.45\textwidth, height=0.20\textwidth]{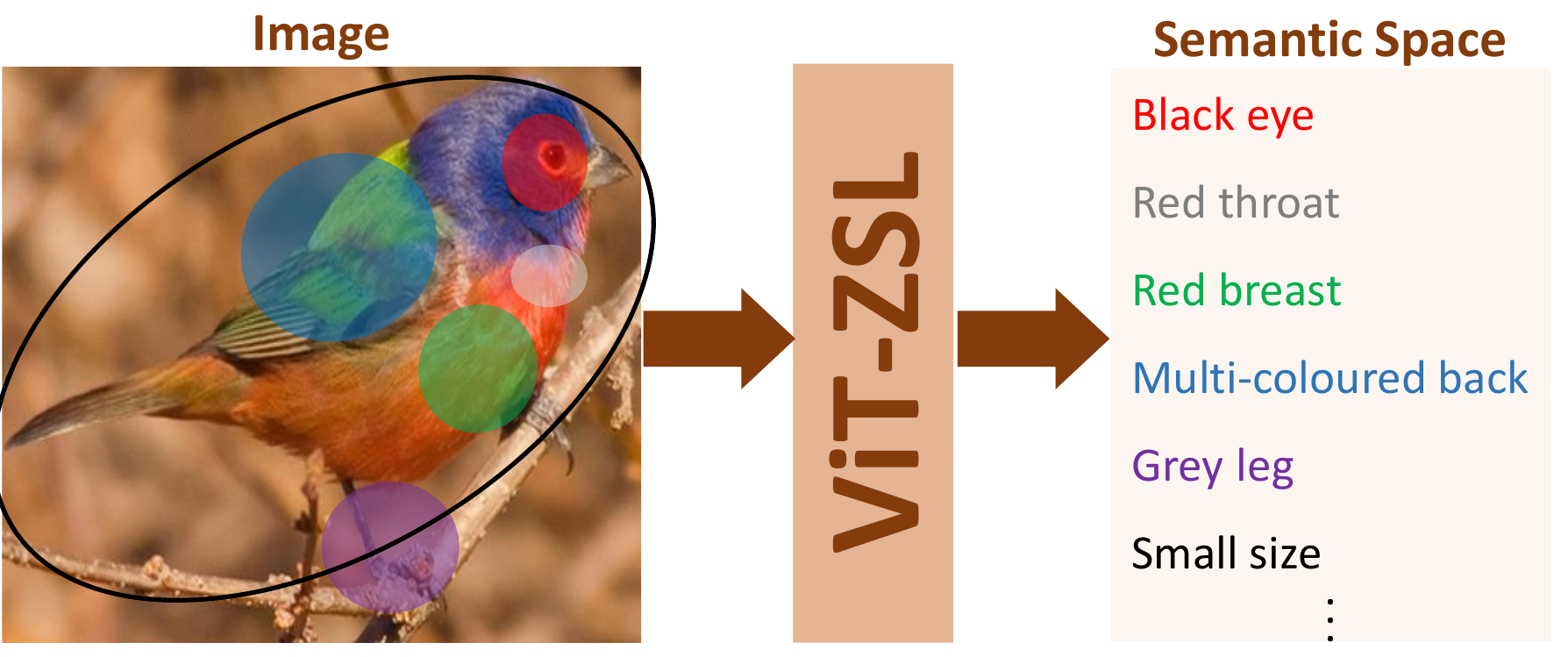}
\end{center}
\vspace{-25pt}
\caption{Our method embeds each attribute-based feature with the semantic space. It learns the visual discriminative features through multi-head attention. \textit{Best to view in colour}: colours in the image correspond to the same-colour attribute in the semantic space.} %\anjan{Change the model name to ViT-ZSL.}
\label{Into_fig}
\end{wrapfigure}
%%%%%%%%%%%%%%%%%%%%%%
\section{Introduction}
Relying on massive annotated datasets, significant progress has been made on many visual recognition tasks, which is mainly due to the widespread use of different deep learning architectures \cite{Ren2015FasterRT, ViT, khan2021transformers}. Despite these advancements, recognising any arbitrary real-world object still remains a daunting challenge as it is unrealistic to label all the existing object classes on the earth. Zero-Shot Learning (ZSL) addresses this problem, requiring images from the \emph{seen} classes during the training, but has the capability of recognising \emph{unseen} classes during the inference \cite{8413121, AREN, APN, Fedirici2020MIB}. Here the central insight is that all the existing categories share a common semantic space and the task of ZSL is to learn a mapping from the imagery space to the semantic space with the help of side information (attributes, word embeddings) \cite{AWA2:journals/corr/XianSA17,Mikolov2013DistributedRO, Pennington2014GloveGV} available with the seen classes during the training phase so that it can be used to predict the class information for the unseen classes during the inference time.

Most of the existing ZSL methods \cite{8578679, Schnfeld2019GeneralizedZA} depends on pretrained visual features and necessarily focus on learning a compatibility function between the visual features and semantic attributes. Although modern neural network models encode local visual information and object parts \cite{AREN}, they are not sufficient to solve the localisation issue in ZSL models. Some attempts have also been made by learning visual attention that focuses on some object parts \cite{SGMA}. However, designing a model that can exploit a stronger attention mechanism is relatively unexplored.

Therefore, to alleviate the above shortcomings of visual representations in ZSL models, in this paper, we propose a Vision Transformer (ViT) \cite{ViT} based multi-head self-attention model for solving the ZSL task. Our main contribution is to introduce ViT for enhancing the visual feature localisation to solve the zero-shot learning task. Without any object part-level annotation or detection, this is the first attempt to introduce ViT into ZSL. As illustrated in Figure \ref{Into_fig}, our method maps the visual features of images to the semantic space with the help of scaled dot-product of multi-head attention employed in ViT. We have also performed detailed experimentation on three public datasets (i.e., AWA2, CUB and SUN) following Generalised Zero-Shot Learning (GZSL) setting and achieved very encouraging results on all of them, including the new state-of-the-art harmonic mean on all the datasets.

\section{Related Work}

% \begin{wrapfigure}{r}{0.5\textwidth} 
%   \vspace{-20pt}
%   \begin{center}
%     \includegraphics[width=0.50\textwidth, height=0.20\textwidth]{Training_Figure.png}
% \end{center}
% \vspace{-20pt}
%   \caption{A schematic view of the main training approaches. \anjan{I think you can remove this figure. This figure is not helping.}}
%   \label{Training_Figure}
% \end{wrapfigure}
\myparagraph{Zero-Shot Learning:} ZSL is employed to bridge the gap between seen and unseen classes using semantic information, which is done by computing similarity function between visual features and previously learned knowledge \cite{romera-paredes15}. Various approaches address the ZSL problem by learning probabilistic attribute classifiers to predict class labels \cite{5206594, ConSE} and by learning linear \cite{DeViSE, SJE, ALE}, and non-linear \cite{LATEM} compatibility function associating image features and semantic information. Recently proposed generative models synthesise visual features for the unseen classes \cite{8578679, Schnfeld2019GeneralizedZA}. Although those models achieve better performances compared to classical models, they rely on features of trained CNNs. Recently, attention mechanism is adapted in ZSL to integrate discriminative local and global visual features. Among them, S$^2$GA \cite{S2GA} and AREN \cite{AREN} use an attention-based network with two branches to guide the visual features to generate discriminative regions of objects. SGMA \cite{SGMA} also applies attention to jointly learn global and local features from the whole image and multiple discovered object parts. Very recently, APN \cite{APN} proposes to divide an object into eight groups and learns a set of attribute prototypes, which further help the model to decorrelate the visual features. Partly inspired by the success of attention-based models, in this paper, we propose to learn local and global features using multi-scaled-dot-product self-attention via the Vision Transformer model, which to the best of our knowledge, is the first work on ZSL involving Vision Transformer. In this model, we employ multi-head attention after splitting the image into fixed-size patches so that it can attend to each patch to capture discriminative features among them and generate a compact representation of the entire image.

\myparagraph{Vision Transformer:} Self-attention-based architectures, especially Transformer \cite{Transfomer} has shown major success for various Natural Language Processing (NLP) \cite{NEURIPS2020_1457c0d6} as well as for Computer Vision tasks \cite{Alamri, ViT}; the reader is referred to \cite{khan2021transformers} for further reading on Vision Transformer based literature. Specifically, CaiT \cite{Deep_transfomer} introduces deeper transformer networks, and Swin Transformer \cite{Swin} proposes a hierarchical Transformer, where the representation is computed using self-attention via shifted windows. In addition, TNT \cite{TnT} proposes transformer-backbone method modelling not only the patch-level features but also the pixel-level representations. CrossViT \cite{CrossViT} shows how dual-branch Transformer combining different sized image patches produce stronger image features. Since the applicability of transformer-based models is growing, we aim to expand and judge its capability for GZSL tasks; to the best of our knowledge, this is still unexplored. Therefore, different from the existing works, we employ ViT to map the visual information to the semantic space, benefiting from the great performance of multi-head self-attention to learn class-level attributes.

\section{Vision Transformer for Zero-shot Learning (ViT-ZSL)}
\label{ProposedModel}
\begin{figure} [!t]
\centering
\includegraphics[width=\linewidth]{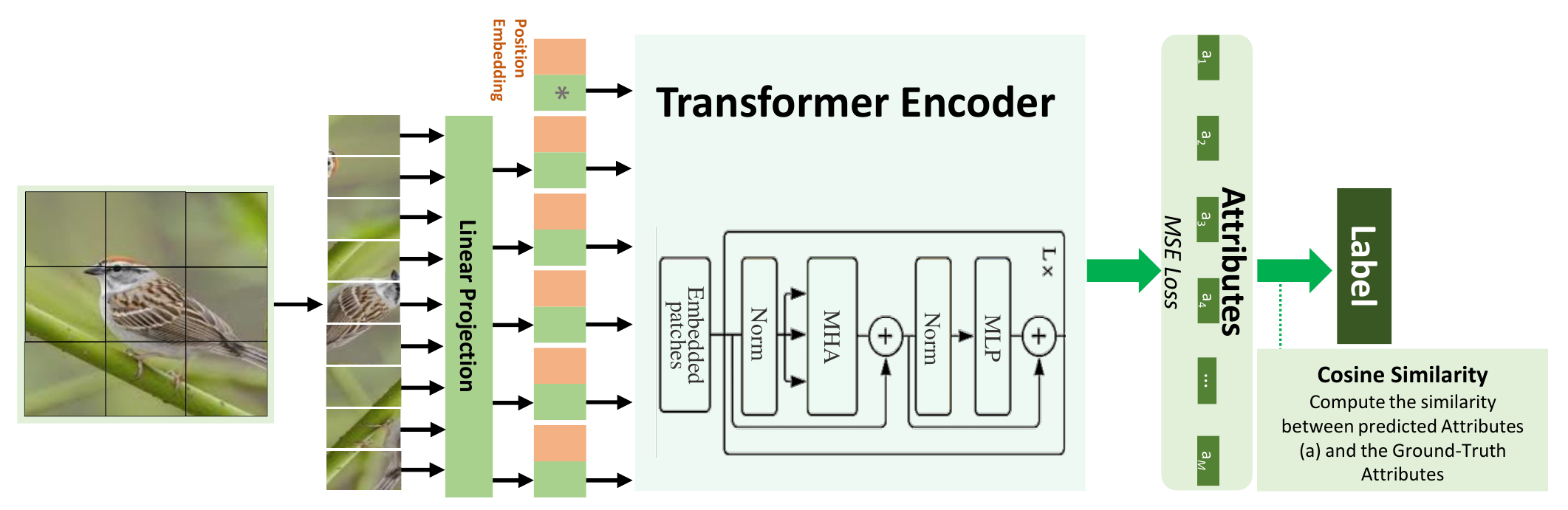}
\vspace{-20pt}
\caption{ViT-ZSL Architecture. An image is split into small patches fed into the Transformer encoder after attaching positional embeddings. During the training the output of the encoder is compared with the semantic information of the corresponding image via MSE loss. At inference the encoder output is used to search for the nearest class label.}
\label{fig:Model}
\end{figure}

%\anjan{The proposed model section is quite short. I think you can expand this further. You can expand the attention mechanism involved in the transformer and explain and emphasise why the attention mechanism is important for ZSL.} 
% We propose our model named \textit{Vision Transformer For Generalised Zero-Short Learning} (ViT-ZSL).
% Following the GZSL testing approach, ViT-ZSL aims to classify both seen classes (\(\mathcal{Y}^s\)) and unseen classes (\(\mathcal{Y}^u\)) (i.e., \(f: \mathcal{X} \rightarrow \mathcal{Y}^u \cup \mathcal{Y}^s)\), where \(\mathcal{X}\) denotes the image space. 
We follow the inductive approach for training our model, i.e. during training, the model only has access to the images and corresponding image/object attributes from the \emph{seen} classes \(\mathcal{S}= \{\mathbf{x}, \mathbf{y} | \mathbf{x}\in \mathcal{X}, \mathbf{y} \in \mathcal{Y}^s \}\), where $\mathbf{x}$ is an RGB image and $\mathbf{y}$ is the class-level attribute vector annotated with $M$ different attributes, as provided with the dataset. As depicted in Figure \ref{fig:Model}, a $224 \times 224$ image $\mathbf{x} \in \mathbb{R}^{H \times W \times C}$ with resolution $H \times W$ and $C$ channels is fed into the model. The model follows ViT \cite{ViT} as closely as possible; hence the image is divided into a sequence of $N$ patches denoted as $\mathbf{x}_p \in \mathbb{R}^{N \times (P^2.C)}$, where \(N = \frac{H.W}{P^2}\). Each patch with a resolution of $ P \times P$ is encoded into a patch embedding by a trainable 2D convolution layer (i.e., Conv2d with kernel size=(16, 16) and stride=(16, 16)). Position embeddings are then attached to the patch embeddings to preserve the relative positional information of the order of the sequence due to the lack of recurrence in the Transformer. An extra learnable classification token ($\mathbf{z}^0_0 = \mathbf{x}_\text{class}$) is appended at the beginning of the sequence to encode the global image representation. Patch embeddings ($\mathbf{z}$) are then projected thought a linear projection $\mathbf{E}$ to $D$ dimension (i.e., $D=1024$) as in Eq. \ref{Eq2.1}. Embeddings are then passed to the Transformer encoder, which consists of Multi-Head Attention (MHA) (Eq. \ref{Eq2.2}) and MLP blocks (Eq. \ref{Eq2.3}). Before every block, a layer normalisation (Norm) is employed, and residual connections are also applied after every block. Image representation ($\mathbf{\hat{y}}$) is produced as in Eq. \ref{Eq2.4}.

\vspace{-15pt}
\begin{align}
\mathbf{z}_0 &=[\mathbf{x}_\text{class}; \mathbf{x}^{1}_{p}\mathbf{E};\mathbf{x}^{2}_{p}\mathbf{E}; \mathbf{x}^{3}_{p}\mathbf{E}; \ldots; \mathbf{x}^{N}_{p}\mathbf{E}] + \mathbf{E}_\text{pos}, & &\mathbf{E} \in \mathbb{R}^{(P^2.C) \times D}, \mathbf{E}_\text{pos} \in \mathbb{R}^{(N+1) \times D} \label{Eq2.1}\\
\mathbf{z}^{\prime}_{\ell} &= \text{MHA}(\text{Norm}(\mathbf{z}_{\ell-1})) + \mathbf{z}_{\ell-1}, & &\ell = 1 \ldots L\;(L=24) \label{Eq2.2}\\
\mathbf{z}_{\ell}&=\text{MLP}(\text{Norm}(\mathbf{z}^{\prime}_{\ell})) + \mathbf{z}^{\prime}_{\ell}, & &\ell = 1 \ldots L \label{Eq2.3}\\
\mathbf{\hat{y}} &=\text{Norm}(\mathbf{z}^0_L) \label{Eq2.4}
\end{align}

In terms of MHA, self-attention is performed for every patch in the sequence of the patch embeddings independently; thus, attention works simultaneously for all the patches, leading to multi-head self-attention. Three vectors, namely Query ($Q$), Key ($K$) and Value ($V$), are created by multiplying the encoder's input (i.e., patch embeddings) by three weight matrices (i.e., $W^Q$, $W^K$ and $W^V$) trained during the training process to compute the self-attention. The $Q$ and $K$ vectors undergo a dot-product to output a scoring matrix representing how much a patch embedding has to attend to every other embedding; the higher the score is, the more attention is considered. The score matrix is then scaled down and passed into a softmax to convert the scores into probabilities, which are then multiplied by the $V$ vectors, as in Eq. \ref{Eq2.5}, where \textit{$d_k$} is the dimension of the $K$ vectors. Since the multi-attention mechanism is employed, self-attention matrices are then concatenated and fed into a linear layer and passed to the regression head. % \anjan{This should not be classification head, it should be regression head, right?}.
\begin{equation}
\text{Attention(\textit{Q}, \textit{K}, \textit{V})}= \text{softmax}(\frac{QK^{T}}{\sqrt{d_k}})\text{\textit{V}}
\label{Eq2.5}
\end{equation}

We argue that self-attention allows our model to attend to image regions that can be semantically relevant for classification and learns the visual features across the entire image. Since the standard ViT has one classification head implemented by an MLP, it has been edited to meet our model objective: to predict $M$ number of attributes (i.e., depending on the datasets used). The motivation behind this is that the network is assumed to learn the notion of classes to predict attributes. For the objective function, we employed the Mean Squared Error (MSE) loss, as the continuous attributes are used as in Eq. \ref{Eq1}, where $\mathbf{y}_i$ is the observed attributes, and $\mathbf{\hat{y}}_i$ is the predicted ones. 
\begin{equation}
\mathcal{L}_\text{MSE} = \displaystyle\frac{1}{M}\sum_{i=1}^{M}(\mathbf{y}_i - \mathbf{\hat{y}}_i)^2 
\label{Eq1}
\end{equation}

%\noindent \anjan{I think in the above equation $n$ should be $M$, please check and confirm.}
During testing, instead of applying the extensively used dot product as in \cite{APN}, we consider the cosine similarity as in \cite{Gidaris2018DynamicFV} to predict class labels. The cosine similarity between the predicted attributes and every class embedding is measured. The output of the similarity measure is then used to determine the class label of the test images.

\section{Experiments} \label{Experiment}

\myparagraph{Implementation Details:} 
All images used in training and testing are adapted from the ZSL datasets mentioned below and sized $224 \times 224$ without any data augmentation. We employ the Large variant of ViT (ViT-L) \cite{ViT}, with input patch size $16 \times 16$, $1024$ hidden dimension, $24$ layers, $16$ heads on each layer, and $24$ series encoder. There are 307M parameters in total in this architecture. ViT-L is then fine-tuned using Adam optimiser with a fixed learning rate of $10^{-4}$ and a batch size of $64$. All methods are implemented in PyTorch\footnote{Our code is available at: \url{https://github.com/FaisalAlamri0/ViT-ZSL}} on an NVIDIA RTX $3090$ GPU, Xeon processor, and a memory sized $32$GB.

% \anjan{Our code will be made available upon acceptance of this paper.

%We employed the Large variant of ViT (ViT-L) \cite{ViT}. ViT-L has $16 \times 16$ input patch size, $1024$ hidden dimension, $24$ layers, $16$ heads on each layer, and 24 series encoder. There are 307M parameters in total in this architecture. ViT-L is then fine-tuned on ZSL datasets, mentioned below, on an NVIDIA GeForce RTX $3090$ GPU, Xeon processor, and memory sized $31$Gi.

\myparagraph{Datasets:} \label{Dataset}
We have conducted our experiments on three popular ZSL datasets: AWA2, CUB, and SUN, whose details are presented in Table \ref{tab:mytab}. The main aim of this experimentation is to validate our proposed method, ViT-ZSL, demonstrate its effectiveness and compare it with the existing state-of-the-arts. Among these datasets, AWA2 \cite{AWA2:journals/corr/XianSA17} consists of $37,322$ images of $50$ categories ($40$ seen + $10$ unseen). Each category contains $85$ binary as well as continuous class attributes. CUB \cite{CUB} contains $11,788$ images forming $200$ different types of birds, among them $150$ classes are considered as seen, and the other $50$ as unseen, which is split by \cite{ALE}. Together with images CUB dataset also contains $312$ attributes describing birds. Finally, SUN \cite{SUN} has the largest number of classes among others. It consists of $717$ types of scene, divided into $645$ seen and $72$ unseen classes. The SUN dataset contains $14,340$ images with $102$ annotated attributes.
\begin{table}[!h]
%\begin{center}
\centering
\caption{Dataset statistics in terms of granularity, number of classes (seen + unseen classes) as shown within parenthesis, number of attributes and number of images.} \label{tab:mytab}
\begin{tabular}{|l|l|l|l|l|}
\hline
Datasets & Granularity & \# Classes   (S + U) & \# Attributes & \# Images  \\ \hline
AWA2 \cite{AWA2:journals/corr/XianSA17} & coarse & 50  (40 + 10)& 85  & 37,322 \\
CUB \cite{CUB} &  fine  & 200  (150 + 50) & 102  & 11,788 \\
SUN \cite{SUN} &  fine   &  717  (645 + 72) & 312 & 14,340 \\
\hline
\end{tabular}
%\end{center}
% \vspace{-20pt}
% \vspace{-10pt}
\end{table}

\myparagraph{Evaluation:} \label{Evaluation}
In this work, we train our ViT-ZSL model following the inductive approach \cite{32933444718}. Following \cite{8413121}, we measure the top-1 accuracy for both seen as well as unseen classes. To capture the trade-off between both sets of classes performance, we use the harmonic mean, which is the primary evaluation criterion for our model. {Following the recent papers (e.g., \cite{APN}, \cite{calibrated}), we apply Calibrated Stacking \cite{calibrated} to evaluate the considered methods under GZSL setting, where the calibration factor $\gamma$ is dataset dependant and decided based on a validation set.} 

%\begin{equation} \label{eq1}
%acc_{Y} = \frac{1}{\textit{$||Y||$}} \sum_{c=1}^{\textit{$||Y||$}}{\frac{\text{\# of correct predictions in c}}{\text{\# of samples in c}}}
%\end{equation}

%\begin{equation} \label{eq2}
%H = \frac{2\times acc_{Y^U} \times acc_{Y^S}}{acc_{Y^U} + acc_{Y^S}} 
%\end{equation}

%\anjan{I think it is better to remove the above two equations. Accuracy and Harmonic Mean are well known metrics. Instead you can provide a reference for the two.}

\myparagraph{Quantitative Results:}
We consider the AWA2, CUB and SUN datasets to show the performance of our proposed model and compare the performance with related arts. Table \ref{tab:Perfromance} shows the quantitative comparison between the proposed model and various other GZSL models. The performance of each model is shown in terms of Seen (S) and Unseen (U) classes and their harmonic mean (H). 

\begin{table} [!h]%{}
\begin{threeparttable}
\centering
\caption{Generalised zero-shot classification performance on AWA2, CUB and SUN}
\label{tab:Perfromance}
\begin{tabular}{|l|c|c|c|c|c|c|c|c|c|c|} 
\hline
\multirow{2}{*}{Models} & \multicolumn{3}{c|}{AWA2} & \multicolumn{3}{c|}{CUB} & \multicolumn{3}{c|}{SUN}   \\ 
\cline{2-10}
& S & U & H & S & U & H & S & U & H \\
\hline
DAP \cite{5206594}   & 84.7 & 0.0 & 0.0 & 67.9 & 1.7 & 3.3 & 25.1 & 4.2 & 7.2 \\
IAP \cite{5206594}   & 87.6 & 0.9 & 1.8 & 72.8 & 0.2 & 0.4 & 37.8 & 1.0 & 1.8 \\
DeViSE \cite{DeViSE} & 74.7 & 17.1 & 27.8 & 53.0 & 23.8 & 32.8 & 30.5 & 14.7 & 19.8 \\
ConSE \cite{ConSE}   & {\color{blue}90.6} & 0.5 & 1.0 & 72.2 & 1.6 & 3.1 & 39.9 & 6.8 & 11.6 \\
SSE \cite{SSE}       & 82.5 & 8.1 & 14.8 & 46.9 & 8.5 & 14.4 & 36.4 & 2.1 & 4.0 \\
SJE \cite{SJE}       & 73.9 & 8.0 & 14.4 & 59.2 & 23.5 & 33.6 & 30.5 & 14.7 & 19.8 \\
ESZSL \cite{romera-paredes15}   & 77.8 & 5.9 & 11.0 & 63.8 & 12.6 & 21.0 & 27.9 & 11.0 & 15.8 \\
LATEM \cite{LATEM}   & 77.3 & 11.5 & 20.0 & 57.3 & 15.2 & 24.0 & 28.8 & 14.7 & 19.5 \\
ALE \cite{ALE}       & 81.8 & 14.0 & 23.9 & 62.8 & 23.7 & 34.4 & 33.1 & 21.8 & 26.3 \\
SAE \cite{SAE}       & 82.2 & 1.1 & 2.2 & 54.0 & 7.8 & 13.6 & 18.0 & 8.8 & 11.8 \\
AREN \cite{AREN}     & {\color{red}92.9} & 15.6 & 26.7  & {\color{red}78.7} & 38.9 & 52.1 & {\color{blue}38.8} & 19.0 & 25.5 \\

%AREN \cite{AREN}     & 79.1 & 54.7 & 64.7  & {63.2} & 69.0 & 66.0 & 40.3 & 32.3 & 35.9 \\

SGMA \cite{SGMA}     & 87.1 & 37.6 & 52.5  & {71.3} & 36.7 & 48.5 & - & - & - \\
APN \cite{APN}       & 78.0 & {\color{blue}56.5} & {\color{blue}65.5}  & 69.3 & {\color{blue}65.3} & {\color{blue}67.2} & 34.0 & {\color{blue}41.1} & 37.6 \\

*GAZSL \cite{GAZSL}   & 86.5 & 19.2 & 31.4  & 60.6 & 23.9 & 34.3 & 34.5 & 21.7 & 26.7 \\
*f-CLSWGAN \cite{8578679} & 64.4 & {\color{red}57.9} & 59.6  & 57.7 & 43.7 & 49.7 & 36.6 & 42.6 & {\color{blue}39.4} \\
%*f-VAEGAN-D2 \cite{Xian2019FVAEGAND2AF} & 76.1 & 57.1 & 65.2  & 75.6 & 63.2 & 68.9 & 50.1 & 37.8 & 43.1 \\ 

\hline
\textbf{Our model (ViT-ZSL)} & \textbf{{90.0}} & \textbf{51.9} & \textbf{{\color{red}65.8}} & \textbf{{\color{blue}75.2}} & \textbf{{\color{red}67.3}} & \textbf{{\color{red}71.0}} & \textbf{{\color{red}55.3}} & \textbf{{\color{red}44.5}} & \textbf{{\color{red}49.3}} \\ \hline
\end{tabular} 

\begin{tablenotes}
\small
\item S, U, H denote Seen classes ($\mathcal{Y}^s$), Unseen classes ($\mathcal{Y}^u$), and the Harmonic mean, respectively. For each scenario, the best is in {\color{red}red} and the second-best is in {\color{blue}blue}. * indicates generative representation learning methods. 
\end{tablenotes}
\end{threeparttable}
\end{table}

DAP and IAP \cite{5206594} are some of the earliest works in ZSL, which perform poorly compared to other models. This is due to the assumptions claimed in these approaches regarding attributes dependency. In real-world animals with attributes `terrestrial' and `farm' are dependent but are assumed independent by such models, which are noted as incorrect by \cite{ALE}. Our model ViT-ZSL does not assume this, but rather it considers the correlation between attributes, which self-attention helps to achieve by considering both positional and contextual information of the entire sequence of patches. DeViSE \cite{DeViSE} and ConSE \cite{ConSE} learn a linear mapping between images and their semantic embedding space. They both make use of the same text model trained on 5.4B words from Wikipedia to construct 500-dimensional word embedding vectors. Both use the same baseline model, but DeViSE replaces the last layer (i.e., softmax layer) with a linear transformation layer. In contrast, ConSE keeps it and computes the predictions via a convex combination of the class label embedding vectors. ConSE, as presented in Table \ref{tab:Perfromance} outperforms DeViSE, but DeViSE is generally performing better on the unseen classes. Similarly, SJE \cite{SJE} learns a bilinear compatibility function using the structural SVM objective function to maximise the compatibility between image and class embeddings. ESZSL \cite{romera-paredes15} uses the square loss to learn bilinear compatibility. Although ESZSL is claimed to be easy to implement, its performance, in particular for GZSL, is poor. ALE \cite{ALE}, which belongs to the bilinear compatibility approach group, performs better than most of its group member. LATEM \cite{LATEM}, instead of learning a single bilinear map, extends the bilinear compatibility of SJE \cite{SJE} as to be an image-class pairwise linear by learning multiple linear mappings. It performs better than SJE on unseen classes but with a lower harmonic mean due to its poor performance on seen classes. Generative ZSL models such as GAZSL \cite{GAZSL}, and f-CLSWGAN \cite{8578679} are seen to reduce the effect of the bias problem due to the inclusion of synthesised features for the unseen classes. However, this does not apply to our method, as no synthesised features are used in our case; instead, solely the features extracted from seen classes are used during training. AREN \cite{AREN}, SGMA \cite{SGMA} and APN \cite{APN} are non-generative ZSL models focusing on object region localisation using image attention. They are the most relevant works to ours as attention mechanism is included in these models architecture. However, they consist of two branches in their models, where the first learns local discriminative visual features and the second captures the image's global context. In contrast, our model uses only one compact network, where the input is the image patches so that the global and local discriminative features can be learned using the multi-head self-attention mechanism.
% As claimed by \cite{Transfomer} \textit{"Self-attention is an attention mechanism relating different positions of a single sequence to compute a representation of the sequence.''} 

\begin{figure}
\centering
\subfigure{
\includegraphics[width=.15\textwidth, height=.1\textwidth]{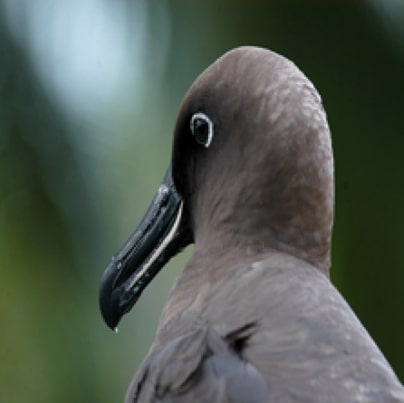}  
\includegraphics[width=.15\textwidth, height=.1\textwidth]{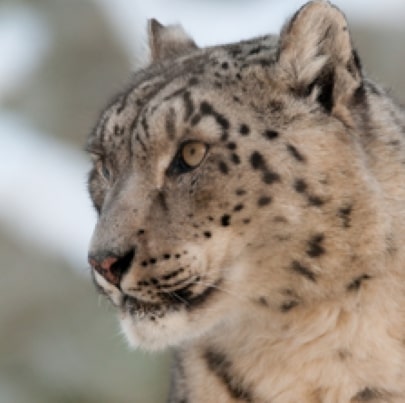}
\includegraphics[width=.15\textwidth, height=.1\textwidth]{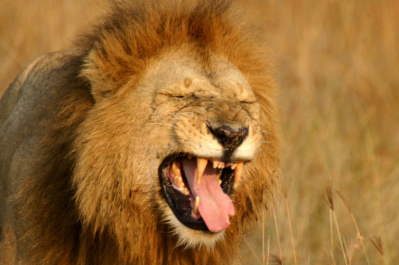}  
\includegraphics[width=.15\textwidth, height=.1\textwidth]{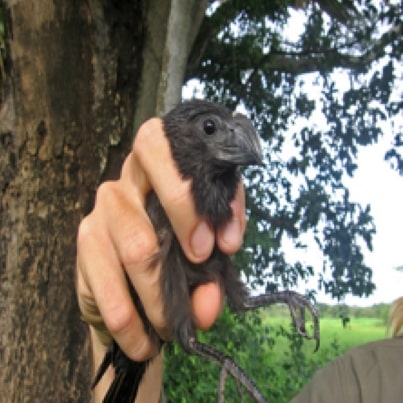}  
\includegraphics[width=.15\textwidth, height=.1\textwidth]{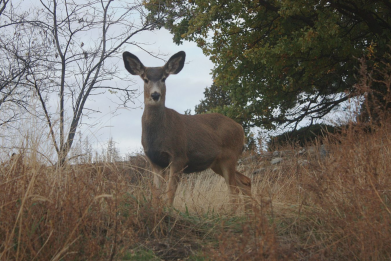}  
\includegraphics[width=.15\textwidth, height=.1\textwidth]{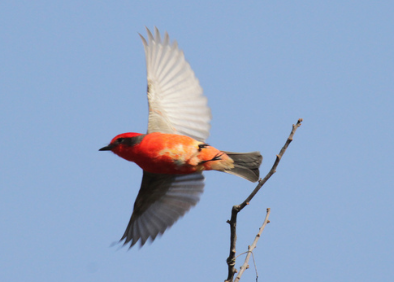}  

}

\subfigure{
\includegraphics[width=.15\textwidth, height=.1\textwidth]{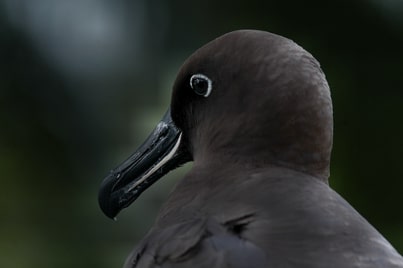}
\includegraphics[width=.15\textwidth, height=.1\textwidth]{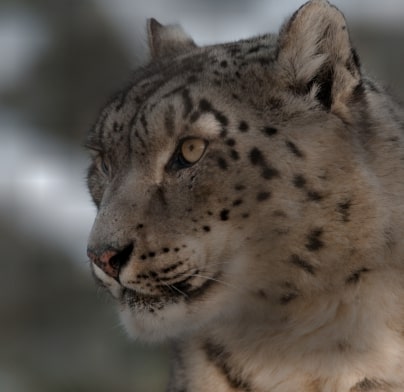}
\includegraphics[width=.15\textwidth, height=.1\textwidth]{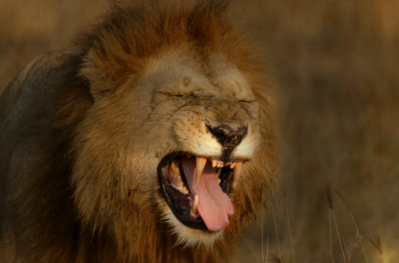}  
\includegraphics[width=.15\textwidth, height=.1\textwidth]{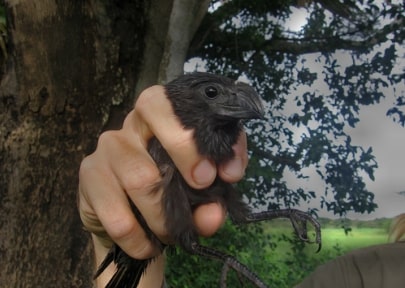}
\includegraphics[width=.15\textwidth, height=.1\textwidth]{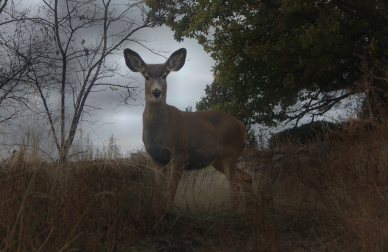}  
\includegraphics[width=.15\textwidth, height=.1\textwidth]{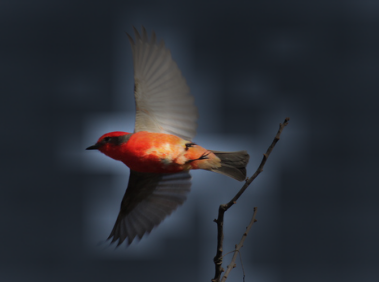}  

}

\subfigure{

\includegraphics[width=.15\textwidth, height=.1\textwidth]{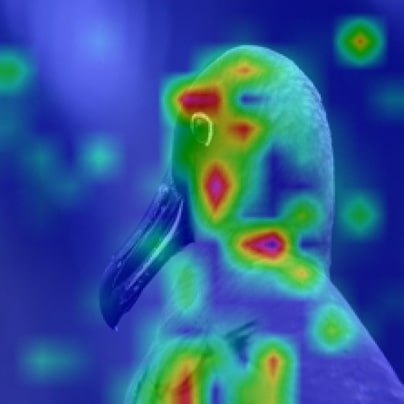}
\includegraphics[width=.15\textwidth, height=.1\textwidth]{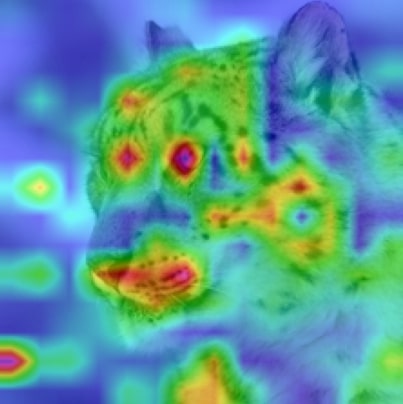}
\includegraphics[width=.15\textwidth, height=.1\textwidth]{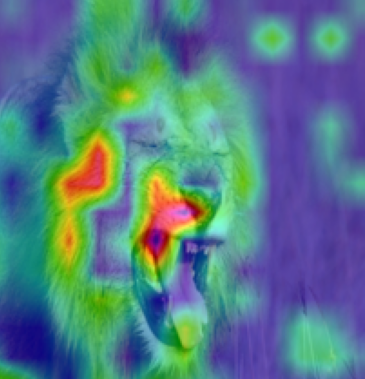} 
\includegraphics[width=.15\textwidth, height=.1\textwidth]{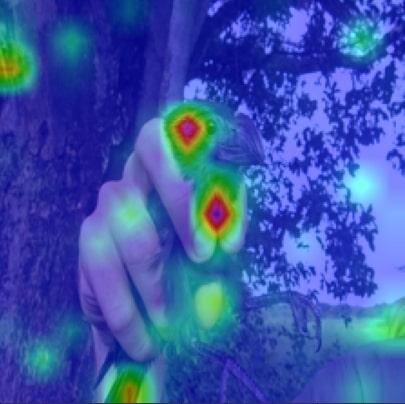}
\includegraphics[width=.15\textwidth, height=.1\textwidth]{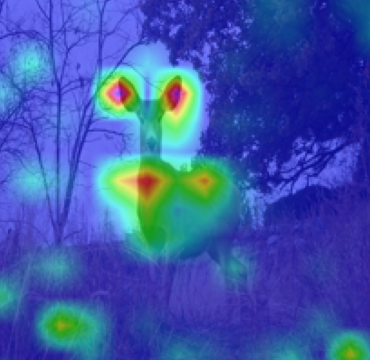} 
\includegraphics[width=.15\textwidth, height=.1\textwidth]{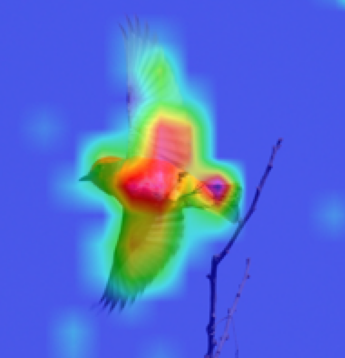}
}
\vspace{-10pt}
\caption{Representative examples of attention. First row: Original images, Middle: Attention maps, and last: Attention fusions. From left to right, ViT-ZSL is able to focus on object-level attributes and learn objects discriminative features when objects are partly captured (first three columns images), occluded (fourth column images) or fully presented (last two columns images).} 
%\anjan{If possible some details on the figure, which is important because sometime people see the figure and read the caption only. Please don't worry about space, I can get some extra space if you want.}}
\label{Att_Figure}
\end{figure}

\begin{comment}

\begin{wrapfigure}{r}{\textwidth} 
  \vspace{-20pt}
  \begin{center}
    %\includegraphics[width=0.50\textwidth, height=0.20\textwidth]{Training_Figure.png}
    %\subfloat[]{\label{main:a}\includegraphics[scale=.2]{example-image-a}}
    \centering
\subfigure{
\includegraphics[width=.13\textwidth, height=.1\textwidth]{images/1.1.jpg}  
\includegraphics[width=.13\textwidth, height=.1\textwidth]{images/1.3.jpg}
\includegraphics[width=.13\textwidth, height=.1\textwidth]{images/1.2.jpg}
}
\subfigure{
\includegraphics[width=.13\textwidth, height=.1\textwidth]{images/2.1.jpg}  
\includegraphics[width=.13\textwidth, height=.1\textwidth]{images/2.3.jpg}
\includegraphics[width=.13\textwidth, height=.1\textwidth]{images/2.2.jpg}
}
\subfigure{
\includegraphics[width=.13\textwidth, height=.1\textwidth]{images/3.1.jpg}  
\includegraphics[width=.13\textwidth, height=.1\textwidth]{images/3.3_U.jpg}
\includegraphics[width=.13\textwidth, height=.1\textwidth]{images/3.2.jpg}
}

\end{center}
\vspace{-20pt}
  \caption{Representative examples of attention. Left: Original images, Middle: Attention maps, and Right: Attention fusions}
  \label{Att_Figure}
\end{wrapfigure}
\end{comment}

% \anjan{Please check `Training\_Figure' has been defined repeatedly.}

Our model ViT-ZSL, as shown in Table \ref{tab:Perfromance}, {achieves the best harmonic mean on AWA2. It also performs as the third best on both seen and unseen classes}. Compared with the other models, {it scores 90.02\%, where the highest is the highest is AREN with 92.9\% accuracy. As the comparison illustrated follows the GZSL setting using the harmonic mean as the primary evaluation metric for GZSL models, ViT-ZSL outperforms all state-of-the-art models.}
%Although ViT-ZSL outperforms all other models on seen classes, its performance on unseen remains limited. We speculate that this is due to the lack of occurrence of some vital and identical attributes between seen and unseen classes. For example, attributes \textit{nocturnal} in bat, \textit{longneck} in giraffe or \textit{flippers} in seal score the highest attributes in the class-attribute vectors, but rarely appear among other classes. 
In terms of the CUB dataset, our method {achieves the second-highest accuracy for seen classes, but
the highest for unseen}. In addition, our ViT-ZSL {obtains the best harmonic mean score among all the reported approaches}. On SUN datasets, which has the most significant number of object classes among other datasets, our model performs as the best for both seen and unseen classes, achieving a harmonic mean of 47.9\%, the highest compared to all other models. 
%\anjan{please try to write a brief explanation on the reason our method is not performing so well on the unseen classes of AWA2 and the difference being quite high 7.1\% - 8.5\%}

\myparagraph{Attention Maps:} In Figure \ref{Att_Figure}, we show how our model attends to image regions semantically relevant to the object class. For example, in the images of the first three columns, the entire objects' shapes are absent (i.e., only the top part is captured), and in the image in the fourth column, the groove-billed ani bird is impeded by a human hand. Although these images suffer from occlusion, our model accurately attends to the objects in the image. Thus, we believe that ViT-ZSL definitely benefits from the attention mechanism, which is also involved in the human recognition system. Clearly, we can say that our method has learned to map the relevance of local regions to representations in the semantic space, where it makes predictions on the visible attribute-based regions. Similarly, in the last two columns images of Figure \ref{Att_Figure}, it can be noticed how the model pays more attention to some object-level attributes (i.e., \textit{Deer}: forest, agility, furry etc., and \textit{Vermilion Flycatcher}: solid and red breast, perching-like shape, notched tail). It can also be noticed that the model focuses on the context of the object, as in the second column images. This can be due to the guidance of some attributes (i.e., forest, jungle, ground and tree) which are associated with \textit{leopard} class. However, as shown in the first column, the model did not pay much attention to the bird's beak compared to the head and the rest of the body, which needs to be investigated further and building an explainable model as in \cite{f-VAEGAN-D2} could be a way to accomplish this.

\section{Conclusion}
In this paper, we proposed a Vision Transformer-based Zero-Shot Learning (ViT-ZSL) model that specifically exploits the multi-head self-attention mechanism for relating visual and semantic attributes. Our qualitative results showed that the attention mechanism involved in our model focuses on the most relevant image regions related to the object class to predict the semantic information, which is used to find out the class label during inference. {Our results on the GZSL task, including the highest harmonic mean scores on the AWA2, CUB and SUN datasets, illustrate the effectiveness of our proposed method}.

Although our method achieves very encouraging results for the GZSL task on three publicly available benchmarks, the bias problem towards seen classes remains a challenge, which will be addressed in future work. Training the model in a transductive setting, where visual information for unseen classes could be included, is a direction to be examined.

\section*{Acknowledgement}
This work was supported by the Defence Science and Technology Laboratory and the Alan Turing Institute. The TITAN Xp and TITAN V used for this research were donated by the NVIDIA Corporation.

%%%%%%%%%%%%%%%%%%%%%%%%
\appendix

%\section{Appendix 1}

\bibliographystyle{apalike}

\bibliography{imvip}

\end{document}